\documentclass[suppldata]{interact}

\usepackage{epstopdf}
\usepackage{subfigure}

\usepackage{natbib}
\bibpunct[, ]{(}{)}{;}{a}{}{,}
\renewcommand\bibfont{\fontsize{10}{12}\selectfont}

\theoremstyle{plain}

\theoremstyle{definition}

\theoremstyle{remark}

\usepackage{graphicx}%
\usepackage{multirow}%
\usepackage{amsmath,amssymb,amsfonts}%
\usepackage{amsthm}%
\usepackage{mathrsfs}%
\usepackage[title]{appendix}%
\usepackage{xcolor}%
\usepackage{textcomp}%
\usepackage{manyfoot}%
\usepackage{booktabs}%
\usepackage{algorithm}%
\usepackage{colortbl}
\usepackage{algorithmicx}%
\usepackage{algpseudocode}%
\usepackage{listings}%
\usepackage{subcaption}%

\usepackage{bbm}
\usepackage{soul}
\usepackage{hyperref}
\usepackage{caption}

\definecolor{codegreen}{rgb}{0,0.6,0}
\definecolor{codeorange}{rgb}{1,0.5,0}
\definecolor{codeblue}{rgb}{0.,0.5,0.99}
\definecolor{codepurple}{rgb}{0.58,0,0.82}
\definecolor{backcolour}{rgb}{0.95,0.95,0.92}
\definecolor{backcolour2}{rgb}{0.96,0.96,0.96}

\newcommand{\bfy}{\mathbf{y}}

\newcommand{\softplus}{\mathrm{softplus}}
\newcommand{\sigmoid}{\mathrm{sigmoid}}
\newcommand{\binomp}{\mathrm{p}}
\newcommand{\notsign}[1]{{\color{gray}#1}}
\newcommand{\signpos}[1]{{\color{black}#1}}
\newcommand{\signneg}[1]{{\color{red}#1}}

\usepackage{etoolbox}

\newbool{anonymous}
\boolfalse{anonymous}
\newcommand{\secretsection}[1]{\ifbool{anonymous}{}{#1}}

\begin{document}

\articletype{}

\title{Intermittent time series forecasting: local vs global models}

\author{
\ifbool{anonymous}{}{
  \name{Stefano Damato\thanks{Corresponding author: Stefano Damato. Email: stefano.damato@supsi.ch}, Nicoló Rubattu, Dario Azzimonti, Giorgio Corani}
  \affil{SUPSI, Istituto Dalle Molle di Studi sull’Intelligenza Artificiale (IDSIA) \\ Lugano, Switzerland}
}
}

\maketitle

\begin{abstract}
Forecasting intermittent time series, which contain zeros, is a crucial challenge in supply chains as inventory policies require probabilistic forecasts to establish safety levels.
Intermittent time series are commonly forecast using local models, trained individually on each time series.
In the last years global models, trained on a large collection of time series, have become popular for time series forecasting. Global models are often based on neural networks or gradient boosted trees.

We carry out the first study comparing state-of-the-art probabilistic local and global models on intermittent time series. For global models we consider three different distribution heads suitable for intermittent time series: negative binomial, hurdle-shifted negative binomial and Tweedie. To the best of our knowledge, this is the first use of the latter two with neural networks.

We perform experiments on five datasets comprising overall more than 40'000 real-world time series. Among global models, TiDE, a simple neural network architecture, achieves the best accuracy; it also consistently outperforms local models and has lower computational requirements.
Large global models are instead much more computationally demanding and less accurate. 
Among the distribution heads, the Tweedie provides the best estimates of the highest quantiles.
\end{abstract}


\begin{practitionersummary}

Across modern supply chains, a large share of inventory items exhibit intermittent demand, generating time series characterised by the presence of zeros. Planning inventory levels requires probabilistic forecasts of these intermittent time series; the predictive distribution should be supported on non-negative values, have a mass at zero and a long upper tail.

This paper provides a guide to practitioners on how to choose appropriate global or local models. We investigate both predictive accuracy and  computational efficiency, reported for five datasets with different characteristics. 
We show that lightweight neural network architectures like TiDE consistently provide the highest forecasting accuracy with limited computational requirements, outperforming local models too. In contrast, larger deep learning architectures like Transformers require high computational overhead and offer less accurate and stable results. While Gradient Boosted Trees  have good performances in producing point forecasts, our experiments show that their probabilistic adaptations are not as effective.

Our study also aims to identify the best distribution to be coupled with global models: we show that at the highest quantiles the Tweedie distribution head offers the best performance, while on lower quantiles there is no major difference between Tweedie, negative binomial and hurdle-shifted negative binomial.

\end{practitionersummary}

\newpage

\section{Introduction}
In supply chain and demand planning, it can be necessary to forecast a large number of 
\textit{intermittent} time series \citep{boylan2021intermittent}, i.e., non-negative time series which contain zeros and are typically count-valued.
Decision making, for instance to set safety amounts \citep{Syntetos_Babai_Boylan_Kolassa_Nikolopoulos_2016}, requires \textit{probabilistic forecasts} which cover non-negative values, have a mass at zero and a long upper tail. 

Many probabilistic models for intermittent time series, including the most recent ones \citep{spiliotis2021product, Sarlo_Fernandes_Borenstein_2023,Svetunkov_Boylan_2023, damato2025, sbrana2025markov},
are trained \textit{locally}, i.e., on individual time series. 
A different approach is instead to train \textit{global} forecasting models  \citep{montero2021principles, hewamalage_global_2022},
by pooling the data of all series in a data set and fitting a single forecasting function \citep{januschowski2020criteria}.
It has been theoretically and empirically shown  that global models can outperform local models, even if the time series of the data set are unrelated \citep{montero2021principles}. 
For instance, neural networks obtained only mixed results when trained  locally; instead, they achieve state-of-the-art performance \citep{benidis2022deep} if trained globally.
So far, the empirical comparisons between local and global models have mostly focused on smooth time series or have only used point forecasts for intermittent time series.
Moreover, some studies involving deep learning forecasting models considered inadequate datasets and benchmarks, as pointed out by \citet{hewamalage_forecast_2023}.

We carry out the first comparison of \textit{probabilistic}  global and local models for intermittent time series.
As local models, we select the in-sample quantiles as a baseline and four advanced models: iETS \citep{Svetunkov_Boylan_2023}, GAS-NB \citep{Sarlo_Fernandes_Borenstein_2023}, TweedieGP \citep{damato2025}, and a Markov Walk model \citep{sbrana2025markov}. Such models provide challenging baselines for the global models introduced later.

There is no established  global model architecture for forecasting intermittent time series, although machine learning and deep learning techniques are widely used \citep{Giannopoulos_Dasaklis_Tsantilis_Patsakis_2025}. 
We compare feed-forward neural networks \citep{Babai_Tsadiras_Papadopoulos_2020}, DeepAR \citep{salinas2020deepar}, DLinear \citep{Zeng_Chen_Zhang_Xu_2023}, and TiDE \citep{das2023longterm},
relatively shallow architectures
which compare favourably to transformer-based models on smooth time series. We include transformer-based models as PatchTST \citep{nie2022time} and Autoformer \citep{autoformer_2021}.
We also include probabilistic adaptations of Gradient Boosted Trees \citep{januschowski2022forecasting}, as they performed well in the M5 competition \citep{makridakis2022M5}.
We leave for future research the comparison with pre-trained forecasting models \citep{tan2024language}.

On global models we also compare  different distribution heads suitable
for intermittent time series, expanding the analysis of \citet{Ziel_2022} to complex architectures. 
We consider the negative binomial distribution, widely used with local models \citep{Kolassa_2016, kolassa2022commentary, spiliotis2021product, Long_Bui_Oktavian_Schmidt_Bergmeir_Godahewa_Lee_Zhao_Condylis_2025}, the hurdle-shifted negative binomial (HSNB, \citet{feng2021comparison})
and the Tweedie distribution \citep{Dunn_Smyth_2005}. To the best of our knowledge, we use for the first time the HSNB and the Tweedie as distribution heads for global models.
The latter was used by \citet{damato2025} as predictive distribution for local models, showing better estimates of the highest quantiles compared to the negative binomial distribution.
We assess the accuracy, the computational cost, and the robustness to random initialisation of the models, running our experiments on more than 40,000 time series from five data sets from retail and supply chain.

The remainder of the paper is organised as follows. Section~\ref{sec:model} introduces the probabilistic local and global models considered in this study, together with the distribution heads adopted for intermittent time series forecasting, and strategies for training and for generating forecasts. Section~\ref{sec:experiments} describes the experimental setup, including the datasets, evaluation metrics, and implementation details. Section~\ref{sec:results} presents and discusses the empirical results, comparing the predictive accuracy and computational requirements of the different approaches. Finally, Section~\ref{sec:conclusions} concludes the paper and outlines directions for future research.

\section{Probabilistic models for intermittent time series}\label{sec:model}

We consider a dataset containing $n$ time series and we denote the $i$-th time series by $\bfy^i$, $i=1, \ldots, n$ and its value at time $t$ by $y_t^i$.
We denote by $\bfy_{s:t}^i$  the sequence of observations of the time series from time $s$ to $t$:
\begin{equation*}
\bfy_{s:t}^i := \{y^i_s, \dots, y_t^i\}.
\end{equation*}

We denote the length of the training set by $T$ and the forecast horizon by $h$.
A probabilistic forecast consists of a \textit{predictive  distribution} (or  \textit{forecast distribution}) for the next $h$ time instants of the $i$-th time series 
\begin{equation*}
p(y^i_{T+1}, \ldots, y^i_{T+h} \mid \bfy_{1:T}^i ).
\end{equation*}

 A probabilistic model returns the above forecast distribution.
Each model has learnable parameters (hereafter, \textit{model parameters}). We differentiate between local and global models depending on how the model parameters are learned: local models are trained on individual time series, while global models are
trained on a large set of time series \citep{januschowski2020criteria}.

\subsection{Local models}\label{sec:local_models}
Intermittent time series are commonly forecast \citep[Sec 13.8]{Hyndman_2021} using
\textit{local} models, i.e., by learning an independent model on each time series.
A number of local models which provide  point forecast only have been proposed to forecast intermittent time series, such as Croston's method \citep{Croston_1972} and its modifications \citep{Syntetos_Boylan_2005, Teunter_Syntetos_Babai_2011}; we do not include them as they do not provide a predictive distribution.

\paragraph*{In-sample quantiles} 
This model assumes the time series to be constituted by i.i.d. samples.
The predictive distribution for each time step
is the empirical distribution of the data; thus, the $q$-th quantile of the $h$-steps ahead predictive distribution equals the $q$-th quantile of the training data.
In-sample quantiles (ISQ) are immediate to compute;  despite their simplicity, they 
provide competitive probabilistic forecasts for intermittent time series, see, e.g., \citet{Kolassa_2016}, \citet{spiliotis2021product}, and \citet{Long_Bui_Oktavian_Schmidt_Bergmeir_Godahewa_Lee_Zhao_Condylis_2025}. 

\paragraph*{iETS} 
iETS  \citep{Svetunkov_Boylan_2023} specializes
the \textit{ETS} model \citep[Chap.8]{Hyndman_2021} to intermittent time series.
Based on
Croston's decomposition \citep{Croston_1972},
iETS identifies, for each observation $y_t$,
the demand $d_t$ and the occurrence $o_t$:
\begin{align*}
d_t = \begin{cases}
y_t &\text{if } y_t > 0 \\
\text{undefined} &\text{otherwise,}
\end{cases}
\qquad o_t  = \mathbbm{1}{(y_t>0)}, 
\end{align*}
where $\mathbbm{1}$ is the indicator function. This decomposition implies that $y_t = o_t \cdot d_t$, for each $t$.
iETS forecasts $d_t$ and $o_t$ with independent 
multiplicative exponential smoothing models.
The best model for the occurrence is selected among five different candidates via AIC.

Its predictive distribution is bimodal, being constituted by a mass at zero (the Bernoulli probability of non-occurrence) and a Gamma distribution for positive demand values.
The multi-step ahead mean forecast is obtained via closed-form formulae, but $h$-step ahead probabilistic forecasts are obtained with autoregressive sampling: a value from the predictive distribution for time $T+1$ is drawn, and it is then appended to the time series and treated as an observed value to compute the  predictive distribution for time $T+2$. The process is repeated up to time $T+h$, and the final output is a sample path of a future trajectory. The simplicity of the model, however, allows it to generate fast predictions.

\paragraph*{GAS-NB}

GAS-NB \citep{Sarlo_Fernandes_Borenstein_2023} is a model derived from the Generalised Autoregressive Score (GAS) framework \citep{creal2013generalized}, which provides a class of forecasting models with non-Gaussian forecast distributions. 
The model uses a negative binomial distribution whose mean varies over time.

The latent function, parametrising the mean of the forecast distribution, is given as the sum of a level component and a seasonal component, as in \citet{Caivano2016}. At each timestamp, the two are updated according to a score derived from negative log-likelihood of the predictive distribution at the previous step.
Parameters such as the starting values of the components, the updating weights, or the size parameter of the negative binomial distribution are estimated via an optimiser.
Forecasts are generated fast in an autoregressive fashion.


\paragraph*{TweedieGP}
TweedieGP \citep{damato2025} is a Bayesian model for forecasting intermittent time series. 
The model sets a Gaussian Process prior \citep{Rasmussen_Williams_GPML} on the latent function, which parametrises the mean of the Tweedie likelihood (see Sec.~\ref{subsec:tweedie}). The posterior distribution of the latent function distribution is computed with Bayes theorem. Faster training times are achieved thanks to variational methods \citep{Hensman_Matthews_Ghahramani_2015}. 
The predictive distribution of TweedieGP accounts for the uncertainty of the latent variable. Using the properties of Gaussian Processes, the $h$-step ahead predictive distribution is obtained efficiently avoiding autoregressive sampling.

\citet{damato2025} compare Tweedie and negative binomial likelihoods for the Gaussian process; they report better forecasts with the Tweedie, especially on the highest quantiles.

\paragraph*{Markov Walk}

\citet{sbrana2025markov} proposes a model where a latent function is controlled by the sum of an autoregressive moving average process and a Markov chain that only takes 0 and 1 as values.

The parameters of the Markov Walk model can be estimated in closed form. Thanks to the fast running times, a validation loop is applied during training to select the starting point of the training set of the time series.
Mean forecasts and predictive variances can be computed with closed-form expressions, and prediction intervals are generated assuming a Gaussian forecast distribution.

\subsection{Global models}
We focus on \textit{global univariate models} \citep{hewamalage_global_2022} which
are trained using all the time series of the data set
and forecast each time series independently.
As for neural networks,
with a slight abuse of terminology we divide them into \textit{shallow}
(feed-forward and DLinear, TiDE) and \textit{large} (DeepAR, PatchTST, Autoformer). We dropped simpler transformer-based architectures, such as the vanilla transformer \citep{vaswani_attention_17} and Informer \citep{haoyietal-informer-2021}, because of their high computational costs and their unstable and uncompetitive results; we do not include models that do not have a publicly available probabilistic implementation (e.g. \citet{xlstm}) Furthermore we consider Gradient Boosted Trees, which were adopted in many solutions of the M5 competition \citep{januschowski2022forecasting}.
Global models do not receive the entire series as input for training and prediction, but only a subset of lagged data points (named \textit{context window} in the jargon of neural networks).

\paragraph*{Feed-forward neural networks} \label{subsec:fnn}
As in \citet{montero2021principles},
we adopt an architecture with 5 layers of 32 units each, with ReLu activations.
While the original model only returns point forecasts, we adapt it to intermittent demand by equipping it with a distribution head.

This feed-forward neural network (FNN) jointly outputs  the parameters of the predictive distributions of the  next $h$ steps, $ \theta_{T+1}, \dots, \theta_{T+h}$, which we call  \textit{distributional parameters}. We refer to this model as a multi-horizon architecture.
The probabilistic forecast for the next $h$-steps is immediate to compute given the context window.

\paragraph*{DLinear}\label{subsec:dlinear}
DLinear \citep{Zeng_Chen_Zhang_Xu_2023} decomposes the time series into trend and  remainder, using a moving average kernel with unit stride. 
Then, each signal passes through a linear layer; the sum of their outputs composes the last layer before the distribution head. 
As the FNNs, DLinear is multi-horizon, i.e., it returns the parameters of the $h$-steps-ahead predictive distribution. The trend-remainder decomposition is inspired by the Autoformer \citep{autoformer_2021}. 

\paragraph*{TiDE}

The architecture of the TiDE (Time-series Dense Encoder) model \citep{das2023longterm} combines deep learning embeddings and linear models. It utilizes a dense encoder-decoder structure built entirely from residual feed-forward blocks. The encoder maps historical data and covariates into a low-dimensional context vector, while the decoder projects this representation into a future horizon matrix. 

A distinctive feature of TiDE is its temporal decoder, which merges these representations with known future covariates (when available) to refine the forecast. Like the FNN and DLinear, TiDE is a multi-horizon architecture that directly outputs the distributional parameters for the next $h$ steps. By avoiding attention mechanisms, it scales linearly with the context and horizon length, offering significantly faster training and inference times than transformer-based models.

\paragraph*{DeepAR} \label{subsec:deepar}
Recurrent neural networks  (RNNs) are competitive global models, particularly when LSTM cell units \citep{lstm1997} are used \citep{Hewamalage_Bergmeir_Bandara_2021,
bandara2019sales}.
\textit{DeepAR} \citep{salinas2020deepar} is a LSTM-based model, designed for  probabilistic forecasting. 
To forecast intermittent time series, this architecture was originally coupled with a negative binomial distribution head; we evaluate it with all three heads described in Sec.~\ref{subsec:tweedie}.

DeepAR adopts autoregressive sampling to  forecast $h$ steps ahead. This strategy with such a large model is, however, computationally demanding; 
the number of simulated trajectories should be decided by considering also the computational cost. 

 


\paragraph*{Autoformer}
The Autoformer \citep{autoformer_2021} is a
transformer-based architecture \citep{vaswani_attention_17} designed for forecasting. It 
decomposes the original time series into  trend and remainder; the same decomposition is applied by DLinear. This operation is performed repeatedly across different steps of a transformer.
The standard transformer architecture is composed by an encoder-decoder structure containing self-attention, a triplet of matrices that is able to capture dependencies across long sequences \citep{vaswani_attention_17}: Autoformer replaces the self-attention with an auto-correlation mechanism, which uses the Fast Fourier Transform to focus on the data with the strongest temporal correlations.

\paragraph*{PatchTST}

PatchTST \citep{nie2022time} introduces two key innovations to adapt the transformer architecture for time-series forecasting: patching and channel-independence. The model slices the time series into non-overlapping shorter patches, which serve as the input units for the transformer's encoder. This design reduces the computational complexity of the self-attention mechanism and allows the model to extract more meaningful local patterns. 
Additionally, the subseries created via patching are fed to the transformer independently: this is proven to make the model more robust. To produce multi-step ahead forecasts, the model uses autoregressive sampling.

\paragraph*{LightGBM}

Gradient boosted trees (GBTs) are widely used in the forecasting literature \citep{januschowski2022forecasting}. LightGBM \citep{ke2017lightgbm} in particular has been part of successful submissions to the M5 competition \citep{makridakis2022M5, LAINDER20221426}, for instance when coupled with the Tweedie loss. 

Yet, natively, GBTs only produce point forecasts \citep{Long_Bui_Oktavian_Schmidt_Bergmeir_Godahewa_Lee_Zhao_Condylis_2025}. Probabilistic forecasts are derived by training several models, either each targeting a specific quantile or deriving intervals from an ensemble of point forecasts.

However, the framework of distributional GBTs \citep{marz2022distributional} allows to couple GBTs with a forecast distribution, training them to predict distributional parameters to minimise the negative log-likelihood.
Multi-step-ahead forecasts can be created in different ways: via autoregressive sampling, training $h$ different models, or training a single model to predict different times. We follow the latter strategy, which is based on the direct strategy of \citet{Bontempi_2013}. For each timestamp $T+j$ the predictions are obtained by appending the number of steps ahead, $j$, to the input of the model.

\subsection{Distribution heads}
We turn neural networks into probabilistic models by equipping them  with a distribution head.
For any parameter $\theta$ of the forecast distribution, we let the model learn an unconstrained parameter $\tilde{\theta} \in \mathbb{R}$. Then, we map the value of $\tilde{\theta}$ into the correct support with an activation function as $\theta = g(\tilde{\theta})$, applying either the softplus function ($\mathrm{softplus}(x) = \log (1 + e^x)$) or the sigmoid function ($\mathrm{sigmoid}(x) = \frac{1}{1+e^{-x}}$).

For neural networks, denoting by $\mathbf{h} \in \mathbb{R}^d$ the last layer of the model before the output, the raw parameter $\tilde{\theta}$ is obtained as a linear projection from $\mathbf{h}$. The parameters of such projection are learnable too. 
For GBTs, $\tilde{\theta}$ is the output of the model.
In the following we discuss the relevant forecast distributions and how we designed their distribution heads.

\subsection{Forecast distributions}

A distribution for intermittent time series should have both a  probability mass at zero and a long upper tail to quantify the probability of large spikes in demand.

\paragraph*{Negative binomial}
The negative binomial has both characteristics; it is thus often used with intermittent demand \citep{Kolassa_2016, spiliotis2021product, Long_Bui_Oktavian_Schmidt_Bergmeir_Godahewa_Lee_Zhao_Condylis_2025}.
We write:
$$Y \sim \mathrm{NegBin}(r, \binomp),$$
where $\binomp \in (0, 1)$, and $r>0$. 
The distribution can have a long upper tail even when its mean is low, 
because  of \textit{overdispersion}:
$\mathrm{Var}[Y]=\frac{\mathbb{E}[Y]}{p} > \mathbb{E}[Y]$.  
For this flexibility it is generally preferred \citep{Kolassa_2016} over the Poisson distribution when modeling intermittent time series;
it is nonetheless bound to be unimodal.

The distribution head is parametrized as:
\begin{equation*}
r = \softplus(\tilde{r}), \quad 
\binomp  =
\sigmoid(\tilde{\binomp}).
\end{equation*}

\paragraph*{Hurdle-shifted negative binomial}
Intermittent time series data have sometimes a bimodal distribution, with a peak mass at zero and a second positive mode corresponding to most frequent demand size.
It thus makes sense to consider a bimodal distribution head for the predictive distribution.
We consider two bimodal distribution heads: the hurdle-shifted negative binomial (HSNB) and the Tweedie. Both are used here for the first time. 

The hurdle-shifted negative binomial (HSNB) \citep{feng2021comparison} is a bimodal generalization of the 
negative binomial. An  independent  Bernoulli random variable  provides the probability ($1 - \pi$) of zero; 
the  negative binomial distribution,  observed with probability $\pi$, is shifted by one unit to model the positive demand. 
Given $Z \sim \mathrm{NegBin}(r, p)$  and $O \sim \mathrm{Ber}(\pi)$, 
the hurdle-shifted negative binomial variable is given by $Y = O \cdot (1+Z)$, which we denote by:
\begin{equation*}\label{eq:hurdleshifted}
Y \sim \mathrm{HSNB}(\pi, r, \binomp)
\end{equation*}
with $\mathbb{P}(Y=0) = 1- \pi$ and for $y >0$, $\mathbb{P}(Y = y) = \pi \cdot \mathbb{P}(Z = y-1)$.
Its parameters are
$\theta = \{ \pi, r, \binomp\}$. 
The shape of the shifted negative binomial, controlled by $r$ and $\binomp$, is independent from the mass at zero, controlled by
 $\pi$. 
The distribution head has parameters:
\begin{equation*}
\pi = \sigmoid (\tilde{\pi}), \
r = \softplus(\tilde{r}), \quad 
\binomp  =
\sigmoid(\tilde{\binomp}).
\end{equation*}

\paragraph*{Tweedie}\label{subsec:tweedie}

The Tweedie distribution \citep{Dunn_Smyth_2005} is bimodal with a peak mass at zero and a continuous long-tailed density on the positive real values.  
Specifically, $Y$ is distributed as a Tweedie if it is an exponential dispersion model \citep{jorgensen1987exponential} with the following power relationship between  mean and variance:
\begin{equation*}
\operatorname{Var}[Y] = \phi \operatorname{E}[Y]^\rho,
\end{equation*}
where $\phi>0$ is the dispersion and $\rho \in (1,2)$ is the power. We denote the mean by $\mu = \operatorname{E}[Y] >0$ and we write 
\[Y \sim \operatorname{Tw}(\mu,\phi,\rho). \]

\begin{figure}[!ht]
\centering
\includegraphics[width=1\linewidth]{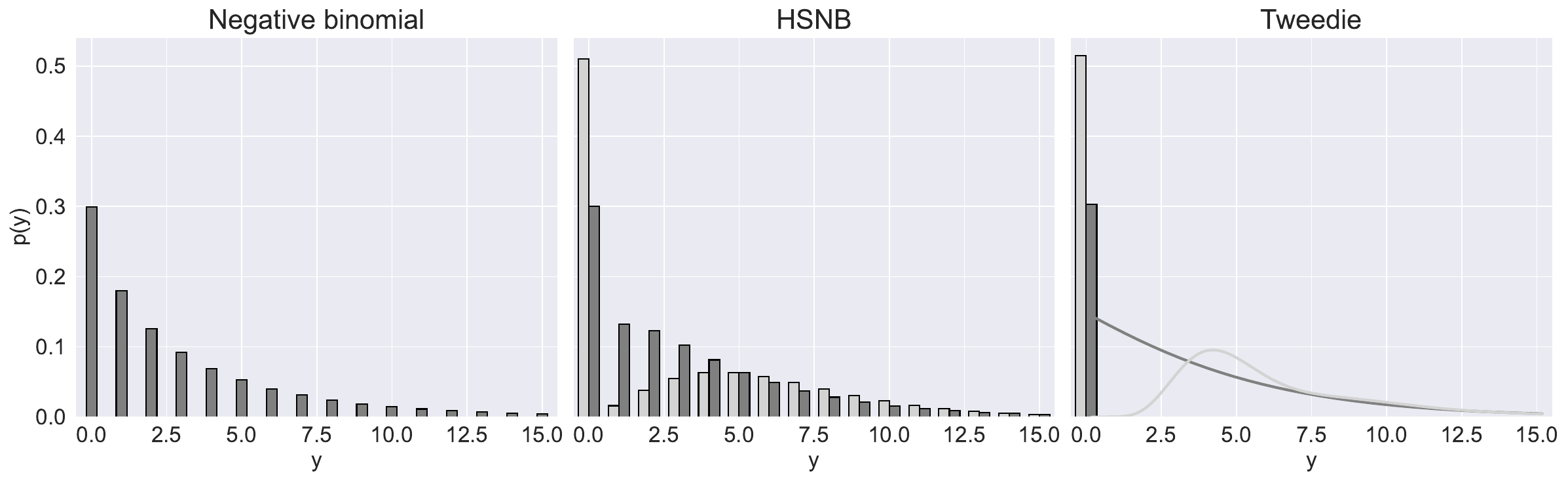}
\caption{Every distribution above  has mean 3 and variance 15. This uniquely determines the parameters of the negative binomial, while different parameterizations are possible
for the Tweedie and the HSNB. The distributions in dark grey have approximately the same mass at zero and similar shape.
Instead, the light grey distributions are examples of bimodal HSNB and Tweedie.
The HSNB is count-valued, while the Tweedie is absolutely continuous on positive values.}
\label{fig:distributions}
\end{figure}

Unlike the HSNB, all distributional parameters affect both the probability in zero and the density of the positive demand. 
Another difference is that its  density is continuous (Fig.~\ref{fig:distributions}).
Thus, it can be the only suitable distribution head if the intermittent time series is real-valued.
To the best of our knowledge, ours is the first implementation of neural networks with a Tweedie distribution head. The Tweedie head is parametrised by
$\theta=\{\mu, \phi, \rho\}$ with:
\begin{equation*}
\mu = \softplus(\tilde{\mu}), \
\phi = \softplus(\tilde{\phi}), \
\rho = 1+\sigmoid(\tilde{\rho}).
\end{equation*}

The \textit{Tweedie loss}, available for instance for LightGBM, is often 
used to train neural networks \citep{jeon2022robust} and GBT models \citep{januschowski2022forecasting}. 
However, it is only  a rough approximation  of the Tweedie density and models trained in this way only return point forecasts;
see \citet{damato2025} for a detailed discussion. Here, we use a fully evaluated Tweedie density.

\subsubsection*{Training}
We denote  the training set length as $T$ and the data set of time series as $\bfy_{1:T+h}$.
We train the models on $\bfy_{1:T-h}$; we use 
 $\bfy_{T-h+1:T}$  as validation set and $\bfy_{T+1:T+h}$ as test set. 

For neural networks, both training and forecast are performed by processing \textit{batches} of time series. A batch $B$ is a collection $(i,w_i)$, where $i$  is the index of a time series and $w_i \in \{1, \ldots, T\}$ is the last index of the context window. 
Model parameters are learned by minimizing the negative log-likelihood of the predictive distribution on the training set summed over the different batches:
\begin{equation}
\label{eq:nll}
\mathcal{L} = - \sum_{(i, w_i)\in B}\sum_{t=1}^{h} \log p(y^i_{w_i+t} | \theta^i_{w_i+t}) 
\end{equation}
where $\theta^i_{w_i+t}$ is the output of an evaluation of the model at $\bfy^i_{w_i-c+1:w_i}$ and $c$ is the context length, which defines how far the training looks back. 
This loss $\mathcal{L}$ is minimized by using Adam optimizer \citep{Kingma_Ba_2017}. We use the negative log-likelihood on the validation set as a criterion for early stopping. We used sensible default values for the hyperparameters of neural networks; an extensive tuning is not feasible due to the exploding computational costs, as detailed in Sec.~\ref{sec:large_models}.

GBTs are designed to work on tabular data: thus, we create tables for training and validation by stacking together sequences with past values sampled from temporal time series, as well as their respective timestamps. For each row, the predictors are $(i, \bfy^i_{w_i-c+1:w_i}, w_i-c+1, \dots, w_i, t_i)$ and the target is $y_{w_i + t_i}$, with $t_i \in {1, \dots, h}$: note that beyond the context window, the predictors include a time series identifier $i$ and time indices for both the context window and the target. This way, a single model learns to produce forecast for different horizons. In this case too, the output of the models are the parameters of the forecast distribution, and the training objective is the negative log-likelihood as in eq.\eqref{eq:nll}.
The training of GBTs includes the identification of hyperparameters: this step plays a fundamental role in their performance \citep{petropoulos2025wielding} and its cost can be regarded as training cost. We follow the procedure used by \citet{erickson2025tabarena}, sampling 100 combinations of the most important parameters and selecting those who minimise the loss on the validation set.

\subsubsection*{Scaling}
When training neural networks, it is important to bring all time series on the same scale \citep{montero2021principles, salinas2020deepar}.
Dealing with intermittent time series, we cannot
simply scale  the time series before training:
integer-valued predictive distributions can only be evaluated on integer data. 

We overcome the problem as follows.
Denoting by $s_i$ the scale factor of the $i$-th time series, we divide the autoregressive input of $i$-th time series 
by $s_i$. To evaluate the likelihood function of eq.~\eqref{eq:nll} and to generate the forecasts on the test set 
we rescale by $s_i$ the parameters of the predictive distribution which are scale-dependent;
see Appendix \ref{appendix:scaling_nns} for more details. 
We use the mean of the \textit{non-zero values} of the $i$-th time series as  $s_i$. Instead, the mean of the time series is not a suitable scale factor: on intermittent time series it can be much smaller than one, thus inflating the positive values.

Scaling is not available in the distributional GBTs implementations we use, as they do not allow for the scaling trick mentioned above. Thus they receive the unscaled time series as input.


\section{Experiments}\label{sec:experiments}

\begin{table}[!ht]
\centering
\begin{tabular}{llll}
\toprule
& Model name & Implementation & Forecasting approach \\
\midrule

& iETS & \texttt{smooth} (R) & Autoregressive (10{,}000 samples) \\
\rowcolor{backcolour} & GAS-NB & ours (R) & Autoregressive (10{,}000 samples) \\
& TweedieGP & \texttt{TweedieGP} (Python) & Multi-horizon (10{,}000 samples) \\
\rowcolor{backcolour} \multirow{-4}{*}{\rotatebox[origin=c]{90}{Local}} & MW & \texttt{MW} (R) & Multi-horizon (closed-form) \\

\midrule

& FNN & \texttt{gluonts} (Python)  & Multi-horizon  (10{,}000 samples) \\
\rowcolor{backcolour} & DLinear & \texttt{gluonts} (Python) & Multi-horizon  (10{,}000 samples) \\
& TiDE & \texttt{gluonts} (Python) & Multi-horizon  (10{,}000 samples) \\
\rowcolor{backcolour} & DeepAR & \texttt{gluonts} (Python)  & Autoregressive  (200 samples) \\
& PatchTST & \texttt{gluonts} (Python)  & Autoregressive  (200 samples) \\
\rowcolor{backcolour} \multirow{-6}{*}{\rotatebox[origin=c]{90}{Global}} & Autoformer & \texttt{transformers} (Python) & Autoregressive  (200 samples)  \\

\bottomrule
\end{tabular}
\caption{Summary of model implementations.}
\label{tab:impl}
\end{table}

We list the packages and forecasting approaches used in Tab.~\ref{tab:impl}.
For local and shallow global models, we used no covariates. When training large global models, we include time-based covariates such as lagged values,  day of the week or month of the year, time from the start of the time series, and time series IDs.  
We take the Tweedie distribution head from  \citet{damato2025};
it is based on PyTorch. We developed our PyTorch implementation of the HSNB distribution head. Additional details about the model's hyper-parameters are reported in Appendix~\ref{appendix:additionalImplementation}.
The code to reproduce the experiments is available at \url{https://anonymous.4open.science/r/iTS-5516} (made anonymous for the peer-review process).

\subsection{Datasets}
We consider the following datasets: 
\textit{M5}: daily item sales from Walmart stores \citep{makridakis2022M5}; \textit{UCI}: daily item sales from an online store \citep{Turkmen_Januschowski_Wang_Cemgil_2021};
\textit{Auto}:  monthly vehicle sales \citep{Turkmen_Januschowski_Wang_Cemgil_2021};
\textit{Carparts}:   monthly car parts sales \citep{expsmooth_2023};
\textit{RAF}: monthly spare parts demand for British Royal Air Force \citep{Syntetos_Boylan_Croston_2005}. 
In Tab.~\ref{tab:datasets} we report
some summary statistics including the average demand interval (ADI) and squared coefficient of variation (CV$^2$).
Larger ADI corresponds to sparser time series, while larger CV$^2$ corresponds to more volatile time series.

As a criterion to define intermittency, we include in our experiment all series having ADI strictly larger than 1, that is, all time series containing at least one zero \citep{Syntetos_Boylan_Croston_2005}. We observed the results remain quite consistent if we use the popular threshold of ADI $> 1.32$, which however was only introduced to discriminate between two simple forecasting methods, and not to provide a definition of intermittency \citep{Long_Bui_Oktavian_Schmidt_Bergmeir_Godahewa_Lee_Zhao_Condylis_2025}.

\begin{table}[!ht] 
\centering
\begin{tabular}{lrrrrrrrr}
\toprule
Dataset & \# of ts & Freq. & T & h & Context & Avg. ADI & Avg. CV$^2$ \\
\midrule
\rowcolor{backcolour}
M5        & 30490 & D & 1941 & 28 & 112 & 6.2 & 0.4 \\

UCI       & 1191  & D & 76   & 14 & 28  & 10.1 & 1.2 \\

\rowcolor{backcolour}
Auto      & 3000  & M & 18   & 6  & 12  & 1.3  & 0.4 \\

Carparts  & 2493  & M & 45   & 6  & 12  & 5.6 & 0.3 \\

\rowcolor{backcolour}
RAF       & 5000  & M & 72   & 12 & 36  & 9.8  & 0.6 \\
\bottomrule
\end{tabular}

\caption{The  considered datasets; ``D" stands for daily and ``M" for monthly.}
\label{tab:datasets}
\end{table}

We expect M5 to be the most favourable data set for large global models, as it contains more than 30,000 daily time series, each spanning over 5 years. Carparts contains one order of magnitude less time series, which are also notably shorter (also because they are monthly); they have anyway  similar intermittency to M5.
UCI and RAF are the most intermittent datasets; the latter shows large demand spikes. Auto has the shortest time series, with length of 18 months.

Tab.~\ref{tab:datasets} also includes the context length used for global models. It was chosen as a multiple of the forecast horizon. While the context length is often regarded as an important hyperparameter, preliminary experiments showed that, provided it is not very short, it does not impact the performance of the selected models.

\subsection{Forecast evaluation}

We evaluate probabilistic forecasts with the \textit{quantile loss}, which is a proper scoring rule \citep{Gneiting_Raftery_2007}:
\begin{align*}
\text{QL}_{q}(y, \hat{y}_q)  = 
\begin{cases} 
  2(1-q) \cdot \big(\hat{y}_{q} - y\big) & \text{ if } y < \hat{y}_{q} \\
  2q \cdot \big(y - \hat{y}_{q}\big) & \text{ if } y \ge \hat{y}_{q} \ ,
\label{eq:quantileLoss}
\end{cases} 
\end{align*}
where $\hat{y}_q$ is the quantile $q$  of the predictive distribution  and $y$ is the observation. 
We consider $q \in \{0.50, 0.80, 0.90, 0.95, 0.99\}$,   evaluating the upper tail of the distribution but not the lower quantiles, which are generally zero.

The quantile loss is scale-dependent. To average the results across time series, 
we scale it  by  the quantile loss of the in-sample quantile on training data \citep{damato2025}:
\begin{equation}
\mathrm{sQL}_{q}(y, \hat{y}_q) = \frac{\mathrm{QL}_{q}(y, \hat{y}_q)} {\frac1T \sum_{t=1}^T \mathrm{QL}_{q}(y_t,\mathrm{ISQ}_q)}.
\end{equation}

We also report the RMSSE, i.e., the root mean squared error between the observation and the  mean of the predictive distribution, scaled by  the in-sample mean squared error of the na\"ive forecast, whose forecast equals the last observation \citep{makridakis2022M5}.

For each score, we report in the results the average of the scores of the time series in a data set.

\section{Results}\label{sec:results}

\subsection*{GBTs and Autoformer are not worth their cost}\label{sec:large_models}

\begin{figure}[!ht]
    \centering
    \includegraphics[width=1\linewidth]{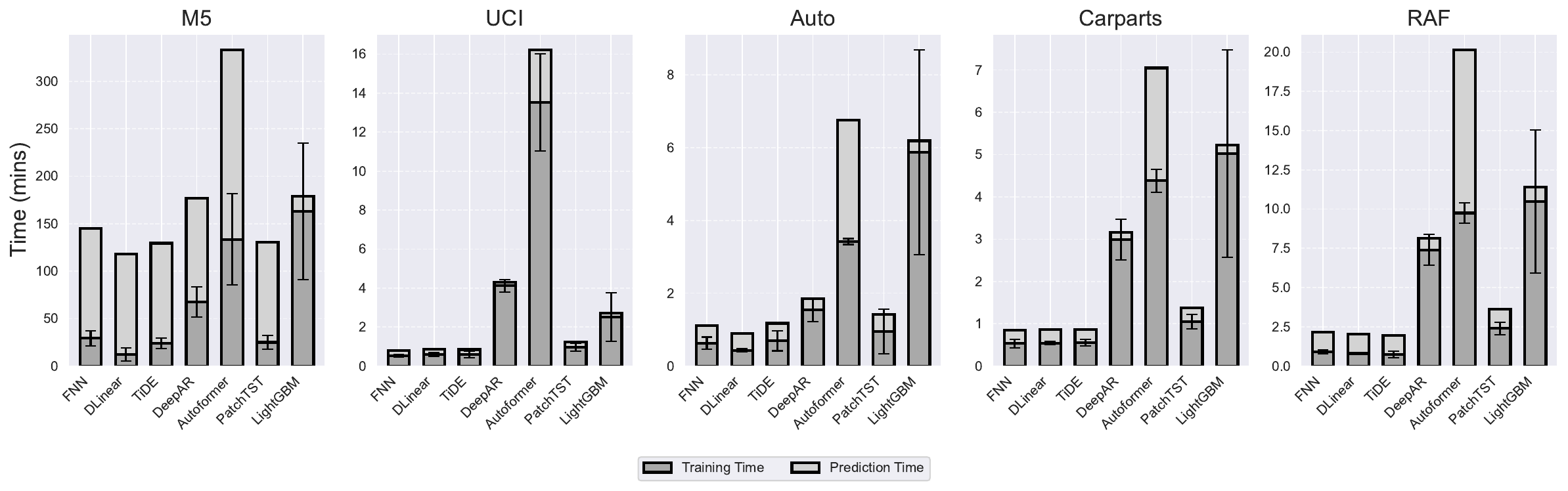}
    
    \caption{Training and prediction times (in minutes) for global models. The whiskers denote the standard deviation of the training times, related to the variability in the amount of epochs for neural networks, and in the duration of any hyperparameter iteration for GBTs.
    }
    \label{fig:global_times}
\end{figure}

In this section, we discuss the computational overhead of the proposed global models; those obtained with different distribution heads are similar. We report in Fig.~\ref{fig:global_times} the computational times for training and generating forecasts. To have comparable outcomes, all models have been run sequentially on the same machine, on the CPU of a Macbook Pro M3 laptop (with Apple M3 Pro chip, 11 cores, 36GB). 

Large models are generally slower than simpler models. On small datasets, their running times are from 2 to 10 times larger; on M5, where generating predictions is expensive for all the models, the difference is less noticeable.

An exception is PatchTST, whose patching strategy is able to lower the computational cost of the transformer architecture. On the contrary, Autoformer is the slowest model. 

LightGBM requires large training times too, as for each iteration it has to be trained on a tabular dataset and has to predict the distributional parameters for the entire validation set.
Forecast generation, which is done in parallel for different time steps, is the fastest across our models.

\begin{figure}[!ht]
\centering
\includegraphics[width=1\linewidth]{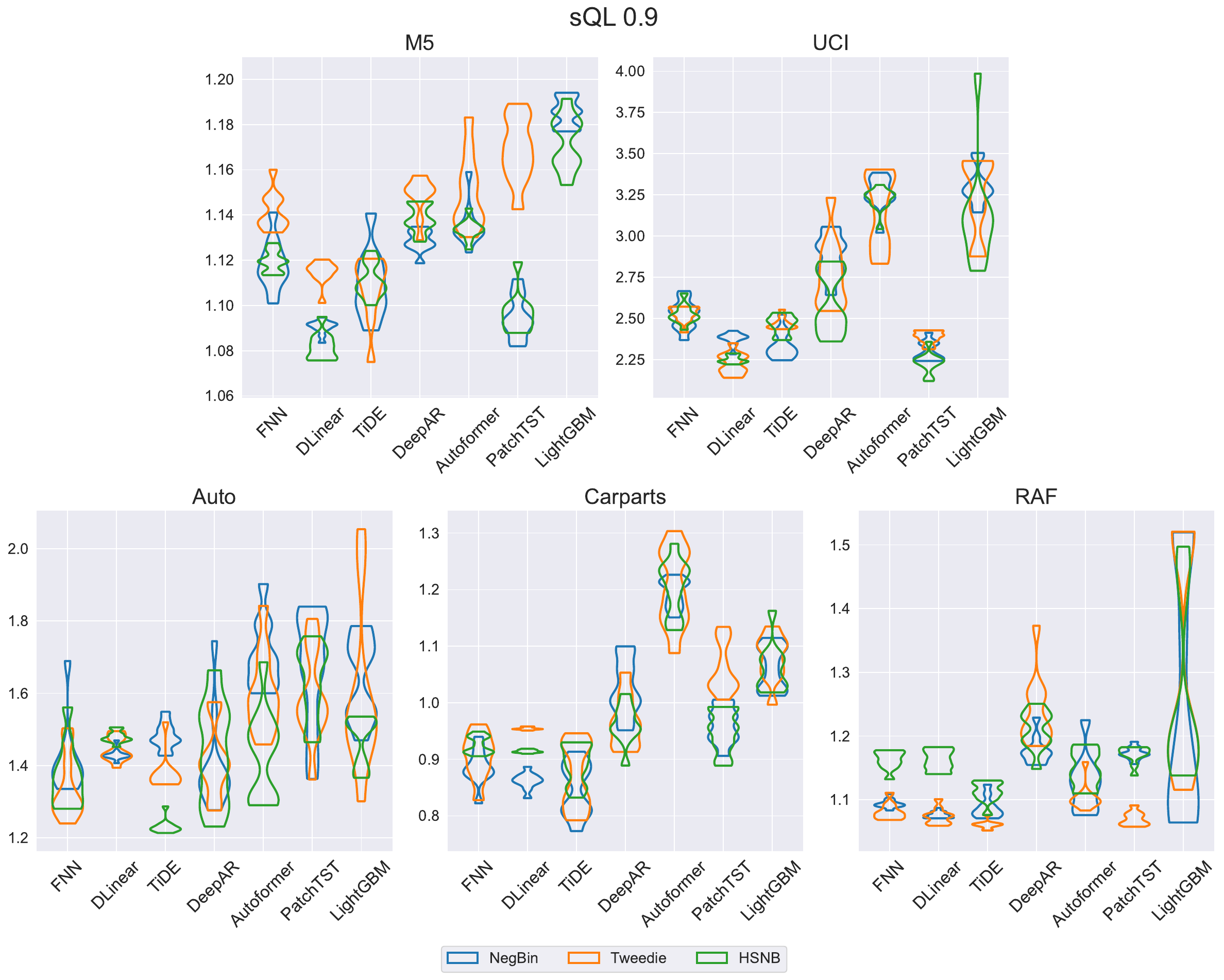}
\caption{Violin plots of the scaled quantile loss (q=0.9) of the global models. For each model and distribution head, 10 runs have been performed with different random seeds; density curves are fit via kernel density estimate. The performance of LightGBM with Tweedie distribution head is not displayed on the M5 data set due to numerical failures.}
\label{fig:scores_example}
\end{figure}

To validate the accuracy and the robustness of global models, we perform 10 different training runs for each model and distribution head: in this way, we also quantify the variability related to random seed initialisation.
Generally such models are trained only once to reduce the computational load; then, the performance achieved would be a single random draw from the distribution we display.

In Fig.~\ref{fig:scores_example}, we show the results 
on quantile 0.9 as an example. The conclusions we discuss here can be drawn based on the exhaustive results of Appendix \ref{appendix:complete_global}, referring to all metrics.
Shallow models appear preferable to large architectures. In particular, the most computationally heavy models, Autoformer and LightGBM, suffer from large variability across different training runs. 
Faster large models, DeepAR and PatchTST, are generally more accurate, although their results worsen in specific cases (e.g. PatchTST with Tweedie head on M5 data set).

Shallow models, and in particular DLinear and TiDE, appear to be the most accurate according to Fig.\ref{fig:scores_example}. These results complement those of \citet{Zeng_Chen_Zhang_Xu_2023}, who already questioned the suitability of transformers for time series forecasting, though only considering point forecast on smooth time series.
In the following we thus drop Autoformer and GBTs from our analysis, as they are both more computationally demanding and less accurate.

\subsection*{Assessing the best neural network architecture} \label{subsec:neural_model_selection}

\begin{table}[!ht]
\centering
\small
\begin{tabular}{cl|ccccc} 
\toprule 
 & Coefficient & M5 & UCI & Auto & Carparts & RAF \\ \midrule 

\multirow{4}{*}{\rotatebox[origin=c]{90}{NegBin}} 
& $\mathbf{c}_\mathrm{model}$ [FNN] & \notsign{0.00} & \signpos{0.36} & \notsign{-0.04} & \signpos{0.06} & \signpos{0.04} \\ 
& $\mathbf{c}_\mathrm{model}$ [DLinear] & \signneg{-0.01} & \signpos{0.30} & \notsign{-0.04} & \notsign{0.00} & \notsign{-0.01} \\ 
& $\mathbf{c}_\mathrm{model}$ [DeepAR] & \notsign{0.00} & \signpos{0.37} & \signneg{-0.11} & \signpos{0.07} & \signpos{0.04} \\  
& $\mathbf{c}_\mathrm{model}$ [PatchTST] & \signneg{-0.02} & \signpos{0.32} & \signpos{0.28} & \signpos{0.07} & \signpos{0.09} \\ 
\midrule

\multirow{4}{*}{\rotatebox[origin=c]{90}{HSNB}} 
& $\mathbf{c}_\mathrm{model}$ [FNN] & \signpos{0.01} & \signpos{0.41} & \signpos{0.14} & \signpos{0.09} & \signpos{0.04} \\ 
& $\mathbf{c}_\mathrm{model}$ [DLinear] & \signneg{-0.01} & \signpos{0.37} & \signpos{0.24} & \notsign{0.03} & \notsign{0.01} \\
& $\mathbf{c}_\mathrm{model}$ [DeepAR] & \signpos{0.03} & \notsign{0.07} & \signpos{0.14} & \notsign{0.02} & \notsign{0.01} \\ 
& $\mathbf{c}_\mathrm{model}$ [PatchTST] & \notsign{-0.00} & \signpos{0.27} & \signpos{0.48} & \signpos{0.08} & \signpos{0.05} \\
\midrule

\multirow{4}{*}{\rotatebox[origin=c]{90}{Tweedie}} 
& $\mathbf{c}_\mathrm{model}$ [FNN] & \signpos{0.03} & \signpos{0.59} & \signneg{-0.06} & \signpos{0.06} & \signpos{0.06} \\
& $\mathbf{c}_\mathrm{model}$ [DLinear] & \notsign{-0.00} & \signpos{0.43} & \signpos{0.04} & \notsign{0.01} & \notsign{0.01} \\ 
& $\mathbf{c}_\mathrm{model}$ [DeepAR] & \signpos{0.08} & \signpos{0.36} & \notsign{0.02} & \signpos{0.06} & \notsign{-0.00} \\ 
& $\mathbf{c}_\mathrm{model}$ [PatchTST] & \signpos{0.10} & \signpos{0.56} & \signpos{0.34} & \signpos{0.16} & \signpos{0.15} \\ 
\bottomrule 
\end{tabular}
\caption{ANOVA coefficients for the comparison of neural networks architectures with TiDE. 
For models entries, positive coefficients imply higher loss than TiDE, and vice versa. We report in black and red, depending on their sign, coefficents that are significantly different ($p$ \textless 0.05) from zero. Coefficients in grey are not significantly different from zero.
We do not report the coefficients of the quantile levels, which do not need to be interpreted.
} 
\label{tab:anova_model}
\end{table}
 
We now statistically compare FNNs, DLinear, TiDE, DeepAR, and PatchTST. For each data set and distribution head, we test the significance of the difference between them, using an ANOVA test; namely we fit a linear regression whose response variable is the score achieved by the model and whose categorical covariates are the model type and the metric type.
The regression formula we use is
\begin{equation*}
y = c_0 + \mathbf{c}_\mathrm{metric}^\top \mathbf{I}_\mathrm{metric} + \mathbf{c}_\mathrm{model}^\top \mathbf{I}_\mathrm{model} + \varepsilon,
\end{equation*}
where $\varepsilon \sim \mathcal{N}(0, \sigma^2)$ is Gaussian noise, $\textbf{I}_\mathrm{model}$ and $\textbf{I}_\mathrm{metric}$ are one-hot encoded vectors
referring to model type (\{FNN, DLinear, DeepAR, PatchTST\}) and metric type $\{$sQL$_q$ for $q \in \{0.5, 0.8, 0.9, 0.95, 0.99\}\}$.
The intercept $c_0$ corresponds to the reference levels, TiDE and RMSSE. The coefficients $\mathbf{c}_\mathrm{model}$ represent the average score difference of DeepAR, DLinear and FNNs compared to TiDE: the former has been chosen as a baseline as it is a recent and fast model.
The coefficients $\mathbf{c}_\mathrm{metric}$ represent the average difference of the metric value with respect to RMSSE;
we do not report them, they are only needed in the model to match the ANOVA assumption of homogeneous variance of the residuals, since different metrics might lie on a different scale. For each distribution head, the regression is fit on ten independent results for each model/metric pair.

The coefficients are shown in Tab.~\ref{tab:anova_model}. A positive coefficient implies worse performance than 
TiDE (higher loss), and vice versa. 

With each distribution head, we observe that PatchTST, DeepAR, and FNNs are generally worse than TiDE. The latter two, moreover, are penalised by larger computational costs.
The only case in which TiDE does not win on most datasets is when it is compared with DLinear, using a negative binomial distribution head.
Thus, the analysis underlines that TiDE is the best neural network architecture for both its accuracy and its fast training and prediction times.

\subsection*{Assessing the distribution head}\label{subsec:head_selection}

\begin{table}[!ht]
\centering
\begin{tabular}{ll|ccccc}
\toprule
Metric & Head & M5 & UCI & Auto & Carparts & RAF \\
\midrule

RMSSE & HSNB & \notsign{0.01} & \notsign{0.05} & \signneg{-0.06} & \notsign{0.00} & \notsign{0.01} \\
RMSSE & Tweedie & \notsign{-0.00} & \signneg{-0.12} & \signneg{-0.05} & \notsign{-0.00} & \notsign{-0.02} \\
\midrule

sQL$_{0.5}$ & HSNB & \signneg{-0.01} & \notsign{-0.00} & \signneg{-0.05} & \notsign{-0.00} & \notsign{-0.00} \\
sQL$_{0.5}$ & Tweedie & \signneg{-0.01} & \notsign{0.01} & \notsign{-0.02} & \notsign{-0.00} & \notsign{-0.00} \\
\midrule

sQL$_{0.8}$ & HSNB & \notsign{0.01} & \notsign{0.00} & \signneg{-0.16} & \signneg{-0.04} & \notsign{-0.00} \\
sQL$_{0.8}$ & Tweedie & \signpos{0.01} & \notsign{0.00} & \signneg{-0.05} & \signneg{-0.03} & \notsign{-0.00} \\
\midrule

sQL$_{0.9}$ & HSNB & \notsign{0.0} & \signpos{0.12} & \signneg{-0.25} & \signpos{0.04} & \notsign{0.02} \\
sQL$_{0.9}$ & Tweedie & \notsign{-0.00} & \signpos{0.13} & \signneg{-0.10} & \notsign{0.02} & \notsign{-0.03} \\
\midrule

sQL$_{0.95}$ & HSNB & \signneg{-0.03} & \signpos{0.21} & \signneg{-0.32} & \notsign{-0.02} & \notsign{-0.01} \\
sQL$_{0.95}$ & Tweedie & \signneg{-0.04} & \signpos{0.13} & \signneg{-0.15} & \signpos{0.04} & \notsign{-0.02} \\
\midrule

sQL$_{0.99}$ & HSNB & \signneg{-0.09} & \signneg{-0.39} & \signneg{-0.55} & \notsign{-0.01} & \notsign{-0.05} \\
sQL$_{0.99}$ & Tweedie & \signneg{-0.13} & \signneg{-0.97} & \signneg{-0.23} & \signneg{-0.24} & \signneg{-0.23} \\

\bottomrule
\end{tabular}

\caption{Coefficients of the ANOVA, run on the results of TiDE models. The predictors are given by the interaction between metric and distribution head. The intercept contains RMSSE among metrics, and negative binomial among heads. Coefficients in red and black are statistically significant, and represent respectively an improvement and a worsening compared to the baseline. Coefficients in grey are not statistically significant.
We do not report the intercept and coefficients related to the negative binomial baseline, as they are not interpreted in our analysis.}
\end{table}

We now study the effect of the distribution head on TiDE, comparing negative binomial, HSNB, and Tweedie.
We run again an ANOVA model where the regressors are the interaction between the metric indicator variable and the distribution indicator variable, using RMSSE and the negative binomial distribution head as reference levels. The model is fit on 30 observations per metric, corresponding to 30 experiments with TiDE (10 for each distribution head).
With this analysis, we want to identify significant deviations in performance from the negative binomial baseline: negative coefficients represent improvements, and positive coefficients signify a worsening.

We observe that on RMSSE and sQL$_{0.5}$ the Tweedie distribution improves the results, although the difference is not always statistically significant, showing that it can efficiently model the distribution's mass near 0.
For intermediate quantile levels ($0.8$, $0.9$, $0.95$) the performance remains competitive with the baseline.

Consistent with \citet{damato2025},
the Tweedie is the best head on the highest quantile; this can be attributed to its flexible tails. Similarly, on the same score the HSNB head shows improvements too over the negative binomial, and has generally a better performance than the baseline.

Thus, we conclude that the Tweedie distribution is very accurate when coupled with the TiDE model.
However, results vary across architectures or data set: for instance, on the Auto dataset the HSNB head generally has lower coefficients than the Tweedie, implying a better performance, while the opposite is observed on the RAF data set. Moreover, as shown in Appendix~\ref{appendix:dlinear_head}, the same analysis run on the DLinear model leads to different conclusions, despite the Tweedie distribution head is still the best one on quantile 0.99.
Thus, excluding such quantile, we recommend to select the neural model and the distribution head via cross-validation, looking at the most relevant quantiles for the application at hand.

\subsection*{Selecting local models}

\begin{table}[!ht]
\centering
\small
\begin{tabular}{ll|ccccc}
\toprule
\rotatebox[origin=c]{90}{Data set} & \rotatebox[origin=c]{90}{Score} & \rotatebox[origin=c]{90}{ISQ} & \rotatebox[origin=c]{90}{iETS} & \rotatebox[origin=c]{90}{GAS-NB} & \rotatebox[origin=c]{90}{TweedieGP} & \rotatebox[origin=c]{90}{MW} \\
\midrule
\multirow{7}{*}{\rotatebox[origin=c]{90}{M5}} 
 & RMSSE & 0.98 & \textbf{0.93} & 2.07 & \textbf{0.93} & 1.14 \\
 & sQL$_{0.5}$ & 1.84 & \textbf{1.74} & 1.76 & 1.75 & 2.00 \\
 & sQL$_{0.8}$ & 1.71 & 1.46 & 1.52 & \textbf{1.44} & 1.73 \\
 & sQL$_{0.9}$ & 1.54 & 1.31 & 1.43 & \textbf{1.25} & 1.65 \\
 & sQL$_{0.95}$ & 1.38 & 1.25 & 1.48 & \textbf{1.16} & 1.77 \\
 & sQL$_{0.99}$ & 1.24 & 1.39 & 2.24 & \textbf{1.17} & 3.00 \\
 \cmidrule{2-7}
 & Time (min) & $<$1 & 345 & 396 & 1537 & 52 \\
\midrule
\multirow{7}{*}{\rotatebox[origin=c]{90}{UCI}} 
 & RMSSE & 1.35 & \textbf{1.34} & 1.43 & 1.35 & \textbf{1.34} \\
 & sQL$_{0.5}$ & \textbf{3.07} & 3.08 & 3.08 & 3.08 & 3.39 \\
 & sQL$_{0.8}$ & \textbf{3.09} & 3.10 & \textbf{3.09} & 3.10 & 3.51 \\
 & sQL$_{0.9}$ & 3.20 & 3.23 & 3.24 & \textbf{3.18} & 3.31 \\
 & sQL$_{0.95}$ & 3.51 & 3.60 & 3.62 & \textbf{3.46} & 3.54 \\
 & sQL$_{0.99}$ & 7.06 & 7.18 & \textbf{6.97} & 7.07 & 8.02 \\
 \cmidrule{2-7}
 & Time (min) & $<$1 & 2 & 1 & 14 & $<$1 \\
\midrule
\multirow{7}{*}{\rotatebox[origin=c]{90}{Auto}} 
 & RMSSE & \textbf{0.82} & \textbf{0.82} & 0.89 & \textbf{0.82} & 1.01 \\
 & sQL$_{0.5}$ & \textbf{1.19} & \textbf{1.19} & 1.24 & \textbf{1.19} & 1.45 \\
 & sQL$_{0.8}$ & 1.31 & 1.32 & 1.37 & \textbf{1.30} & 1.84 \\
 & sQL$_{0.9}$ & 1.48 & 1.48 & 1.51 & \textbf{1.42} & 2.43 \\
 & sQL$_{0.95}$ & 1.8 & 1.74 & 1.75 & \textbf{1.63} & 3.56 \\
 & sQL$_{0.99}$ & 4.16 & 3.1 & 2.73 & \textbf{2.62} & 11.02 \\
 \cmidrule{2-7}
 & Time (min) & $<$1 & 3 & 1 & 21 & $<$1 \\
\midrule
\multirow{7}{*}{\rotatebox[origin=c]{90}{Carparts}} 
 & RMSSE & 0.64 & 0.60 & 0.62 & \textbf{0.59} & 0.64 \\
 & sQL$_{0.5}$ & 1.07 & \textbf{1.03} & 1.04 & 1.05 & 1.20 \\
 & sQL$_{0.8}$ & 1.12 & \textbf{1.02} & 1.03 & 1.05 & 1.21 \\
 & sQL$_{0.9}$ & 1.19 & 1.11 & 1.10 & \textbf{1.09} & 1.27 \\
 & sQL$_{0.95}$ & 1.26 & 1.31 & \textbf{1.16} & 1.17 & 1.57 \\
 & sQL$_{0.99}$ & 1.79 & 2.92 & \textbf{1.53} & 1.54 & 4.19 \\
 \cmidrule{2-7}
 & Time (min) & $<$1 & 3 & 2 & 23 & $<$1 \\
\midrule
\multirow{7}{*}{\rotatebox[origin=c]{90}{RAF}} 
 & RMSSE & \textbf{0.61} & 0.62 & 0.80 & \textbf{0.61} & \textbf{0.61} \\
 & sQL$_{0.5}$ & \textbf{1.00} & \textbf{1.00} & 1.24 & \textbf{1.00} & 1.25 \\
 & sQL$_{0.8}$ & \textbf{1.00} & \textbf{1.00} & 1.09 & 1.02 & 1.26 \\
 & sQL$_{0.9}$ & 1.10 & \textbf{1.08} & 1.14 & 1.15 & 1.15 \\
 & sQL$_{0.95}$ & \textbf{1.24} & 1.29 & 1.31 & 1.26 & 1.31 \\
 & sQL$_{0.99}$ & 2.12 & 2.61 & 2.47 & \textbf{2.10} & 3.36 \\
 \cmidrule{2-7}
 & Time (min) & $<$1 & 9 & 4 & 55 & $<$1 \\
\bottomrule
\end{tabular}
\caption{Performance of selected local methods on the metrics introduced above.} 
\label{tab:local}
\end{table}


We present in Tab.~\ref{tab:local} the results of the local models introduced in Sec.~\ref{sec:local_models}. Computational times are measured on the same machine as those in Fig.~\ref{fig:global_times}. We note that in-sample quantiles are a robust choice on the UCI and RAF datasets, those with the larger ADI coefficients. On the contrary, where the demand size appears less sporadically, dynamic models have an edge.

TweedieGP, for instance, shows a good performance on all datasets. In particular, the flexibility of the Tweedie likelihood allows it to be the best model across three datasets on quantiles 0.95 and 0.99. Moreover it is the best local models according to all the metrics on the M5 data set. Being a Bayesian method, it may be less sensitive to the short training set. TweedieGP, however, requires larger computational costs than its competitors.

On RMSSE and sQL$_{0.5}$, iETS achieves the best score on three datasets out of five (like TweedieGP), but it has lower computational times. Moreover, it is often the second or third best model on higher quantiles, and we thus consider it a competitive choice.

The fastest dynamic model is the Markov Walk. Although originally introduced for the M5 data set, the model shows a good RMSSE performance also on the sparsest data sets, matching that of ISQ. Its quantiles forecasts, however, are penalised by the use of Gaussian likelihood: being symmetric, it struggles to model contemporarily a short left tail for the mass at zero, and a heavy right tail. 

The GAS-NB model often has competitive performances, except on the M5 dataset, which has the longest time series. Its results are overall similar to those of iETS, but the computational cost is the largest after TweedieGP.

\subsection*{Local vs global models}

\begin{figure}[!ht]
\includegraphics[width=1\linewidth]{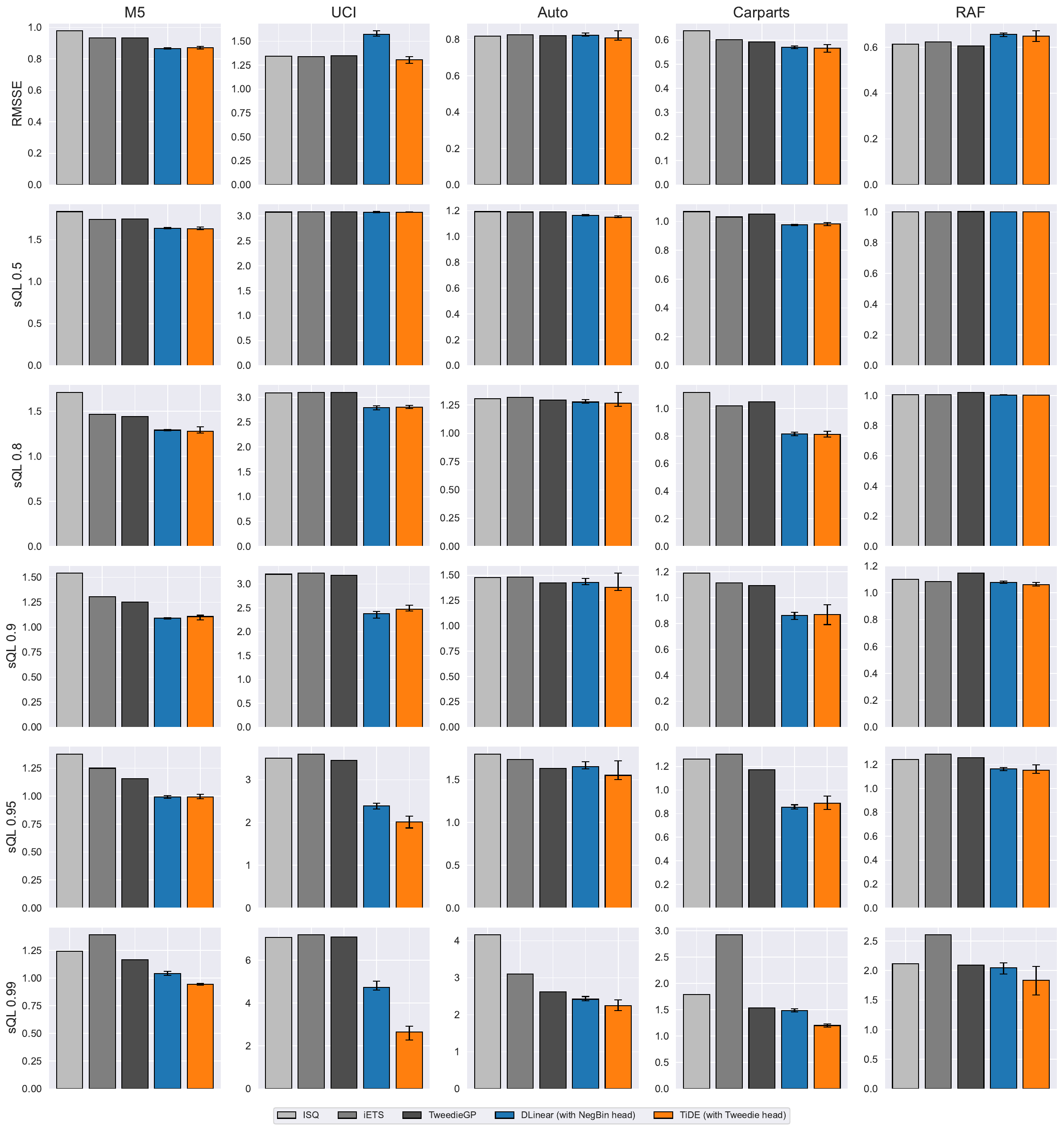}
 \caption{Local models vs DLinear model with negative binomial head. The whiskers on the last columns display the range between the best and the worst of ten runs of the model.}
 \label{fig:local_vs_global}
\end{figure}

In this section, we discuss the performance of the best local and global candidates identified in the previous section. Fig.~\ref{fig:local_vs_global} displays the performance of the TiDE model, equipped with a Tweedie head, and the DLinear, equipped with a negative binomial head, versus in-sample quantiles, iETS, and TweedieGP, which we selected for their computational efficiency or their performances respectively.

The most important finding is the better performance of global models compared to local models on the upper part of the forecast distribution (quantiles 0.8, 0.9, 0.95). At the extreme percentile (0.99), TweedieGP is competitive, however, without outperforming global models. RMSSE is an exception: local methods, for instance, outperform DLinear on three data sets out of five.
In general, these results confirm the superiority of the global approach also on intermittent time series, as long as too complex architectures are avoided.

The results on M5 constitute an exception. The large number of series, their length, and the low volatility of the demand allows all global models (including some of those discarded in Sec.~\ref{sec:large_models}) to be generally stable and effective
(Appendix \ref{appendix:complete_global}). However, shallow models are still preferable to deep ones, both in terms of computational cost and accuracy. In particular, TiDE is very stable and outperforms local models on all metrics.
The other dataset on which TiDE and DLinear outperform local models across all quantiles is Carparts, which has low values of CV$^2$ too. Although less data is available, the demand size is very low, and therefore the quantile loss does not give large penalties.

On the other data sets, shallow models  generally outperform local models. A few exceptions exist: on the RAF data set, which is very sparse and has large demand spikes, ISQ provides the best forecast in terms of RMSSE and moderately high quantiles (0.8, 0.9).
On the Auto data set, TweedieGP outperforms DLinear on quantiles 0.9 and 0.95; but is outperformed by TiDE. 
 
The times reported in Fig.~\ref{fig:global_times} and Tab.~\ref{tab:local} shows that global models are also preferable in terms of computational overhead: indeed, TweedieGP, often the most accurate local model, is also the most computationally expensive one and its training and prediction times are one order of magnitude larger than those of DLinear and TiDE. The cost of local models can easily be reduced by parallelisation \citep{januschowski2020criteria},
but this requires an appropriate infrastructure. 
To further reduce running times with respect to TiDE, in-sample quantiles can be used, as their running times are almost immediate, and they can even outperform some global models in few cases (e.g. RMSSE on UCI, Auto and RAF data sets, or quantile 0.99 on RAF data set). 

Furthermore, we discussed above that large global models may suffer from instability, and that multiple runs may be necessary to obtain reliable results. This is not the case for the shallow models we selected, which show very stable performances (see the whiskers in Fig.~\ref{fig:local_vs_global}).

\section{Conclusions}\label{sec:conclusions}

We performed the first comparison of local and global models for forecasting intermittent time series. We assessed both their accuracy (on multiple quantile levels) and computational requirements.
Concerning global models,
contrary to \citet{petropoulos2025wielding}, we do not observe a trade-off between accuracy and computational overhead: large models, which have larger training times, are also less accurate or provide unstable results across different runs of the experiments. In particular, we observed that gradient boosted trees, which have been used successfully to generate point forecasts \citep{makridakis2022M5}, are not competitive in their probabilistic adaptation, as also reported by \citet{Long_Bui_Oktavian_Schmidt_Bergmeir_Godahewa_Lee_Zhao_Condylis_2025}.
 
Indeed, experiments have shown that a simple model (TiDE) is preferable to both state-of-the-art RNNs (DeepAR) and transformers both in terms of accuracy and computational costs, confirming on intermittent time series the findings of \citet{Zeng_Chen_Zhang_Xu_2023}. As for the distribution head, we observe that the best option may vary on the dataset or the chosen quantile level, as in \citet{Ziel_2022}. However, we observed that the Tweedie distribution is generally very effective on the highest quantile, and is very competitive even on lower levels if coupled with the TiDE model; on the contrary, there is no strong evidence to prefer another distribution over the negative binomial with the DLinear model.  Moreover, real valued intermittent time series, for instance obtained by multiplying the sales of an item by its price, can only be modelled using Tweedie distributions.

We applied existing methods off-the-shelf with minimal hyper-parameter tuning. We are aware that, in certain applications, the use of advanced regularisation techniques and hyper-parameter selection can further enhance the performance of large neural architectures. However, this would require a massive computational overhead and is out of our scope.

In conclusion, we recommend a relatively lightweight model (TiDE) coupled with a Tweedie distribution head, in particular if one is concerned with the highest quantiles. It generally provides more accurate forecasts and faster training times than local models and than other neural networks competitors. 

\section*{Data availability}

The experiments are based on publicly available data sets and model implementations. We release the code to reproduce the experiments at \url{https://anonymous.4open.science/r/iTS-5516}.

\secretsection{\section*{CRediT roles}
SD: Conceptualization, Data curation, Formal analysis,  Investigation, Methodology, Software, Validation, Visualization, Writing – original draft, Writing – review \& editing.
NR: Conceptualization, Data curation, Investigation, Methodology, Software.
DA: Conceptualization, Formal analysis, Funding acquisition, Methodology, Project administration, Resources, Supervision, Validation, Visualization, Writing – original draft, Writing – review \& editing.
DA: Conceptualization, Formal analysis, Funding acquisition, Methodology, Project administration, Resources, Supervision, Validation, Visualization, Writing – original draft, Writing – review \& editing.}

\secretsection{\section*{Funding details}
This work is partially funded by the Swiss National Science Foundation (SNF), grant 200021\_212164/1. This project has received funding from the European Union’s Horizon Europe Research and Innovation Framework under grant agreement No.101160720.}

\bibliographystyle{tfcad}
\bibliography{bibliography}

\newpage

\begin{appendices}

\section{Scaling for neural networks}\label{appendix:scaling_nns}

For neural networks, scaling is computed per batch. Positive values, divided by the scaling factor are no longer necessarily integers. The Tweedie distribution is defined as a continuous distribution on the positive real line, thus the scaling can be applied in a preprocessing step. 
For integer-supported distributions, such as the negative binomial, the scaling is applied on the parameters \citep{salinas2020deepar}. Let $r$ and $\mathrm{p}$ be respectively the count and probability parameters of a negative binomial distribution. Define \[
\mathrm{logit} = \log\left(\frac{1-\binomp}{\binomp}\right).
\]
This transformed parameter is defined on the real line. Using it instead of $\mathrm{p}$ to parametrize a negative binomial distribution $\mathrm{NegBin}\left(y; r, \mathrm{logit}\right)$ has the advantage that 
the expected value of the distribution is $r \cdot e^{\mathrm{logit}}$, thus the predictive parameters are obtained substituting $\mathrm{logit} + \log(s)$ for $\mathrm{logit}$. The mean under the scaled parameters is $s \cdot r \cdot e^{\mathrm{logit}}$. 

The criticality is that, were we to multiply the predictive samples by $s$ (as done with the Tweedie distribution), the resulting variance would be $s^2 \cdot r \cdot e^{\mathrm{logit}}$. However, this parameter-scaling trick yields a variance of $(1 + s \cdot e^\mathrm{logit})\cdot s \cdot r \cdot e^{\mathrm{logit}}$; the limitation is acknowledged by \citet{salinas2020deepar}.

Recalling the formulation in~\eqref{eq:hurdleshifted}, the same strategy has to be applied to the parameter $\binomp$ of the hurdle-shifted negative binomial distribution.

\section{Additional implementation details}
\label{appendix:additionalImplementation}

All neural network architectures are based on \texttt{PyTorch} \citep{paszke2017automatic}, as well as the implementation of the distributions, which allow for automatic differentiation with respect to their parameters.
The implementation of the Tweedie distribution is taken from \citet{damato2025}, and is based on PyTorch too. We also provide a PyTorch implementation of the HSNB distribution.
These are converted into distribution heads for neural networks wrapping around the \texttt{transformers} and \texttt{gluonts} API.

We use the \texttt{gluonts} \citep{gluonts_jmlr} implementation of feed-forward neural networks and DLinear, which do not include covariates. Since the predictive parameters are returned as output in a single step, 10.000 samples are drawn.

When working with deep models, longer forecasting horizons are achieved through autoregressive sampling. Since this procedure is computationally expensive, we just draw 200 sample paths, the default choice in the \texttt{gluonts} implementation. In these models, additional features such as lagged values, time from the start of the time series, and time series IDs are included as input features. 

DeepAR and TiDE models embed the IDs (categorical) feature in a $\mathbb{R}^3$ space and stack two LSTM layers of hidden dimension 40. The implementation we used is provided by \texttt{gluonts}.

PatchTST and Autoformer models embed the IDs in a $\mathbb{R}^3$ space too, and use an embedding layer of size 32, followed by 4 encoder layers and 4 decoder layers, each with two heads and a feed-forward layer of dimension 32. Their implementations are taken from \texttt{gluonts} and HuggingFace's \texttt{transformers} \citep{wolf-etal-2020-transformers} respectively.


 For local models, when possible we used the implementations released by the authors of the respective papers: for iETS, implemented by the R package \texttt{smooth} \citep{Svetunkov_smooth}, the occurrence model is automatically chosen via AIC among five possibilites; TweedieGP is based on the \texttt{GPyTorch} \citep{gardner2018gpytorch} implementation released by the authors, which uses an RBF kernel; for Markov Walk, we used the R implementation released by the author.
 We developed our own implementation of GAS-NB in R.
No covariates were included in the local models.

Local models use all values in the time series. In order to train global models we need to set context length, i.e. how far back the models look to make the prediction. However, since this hyperparameter is expensive to tune, we choose it heuristically as a multiple of the length of the forecast horizon $h$. Preliminary experiments showed that context length does not play a critical role, provided it is sufficiently large.

\section{Complete results of global models}\label{appendix:complete_global}

\begin{figure}
    \centering
    \includegraphics[width=1\linewidth]{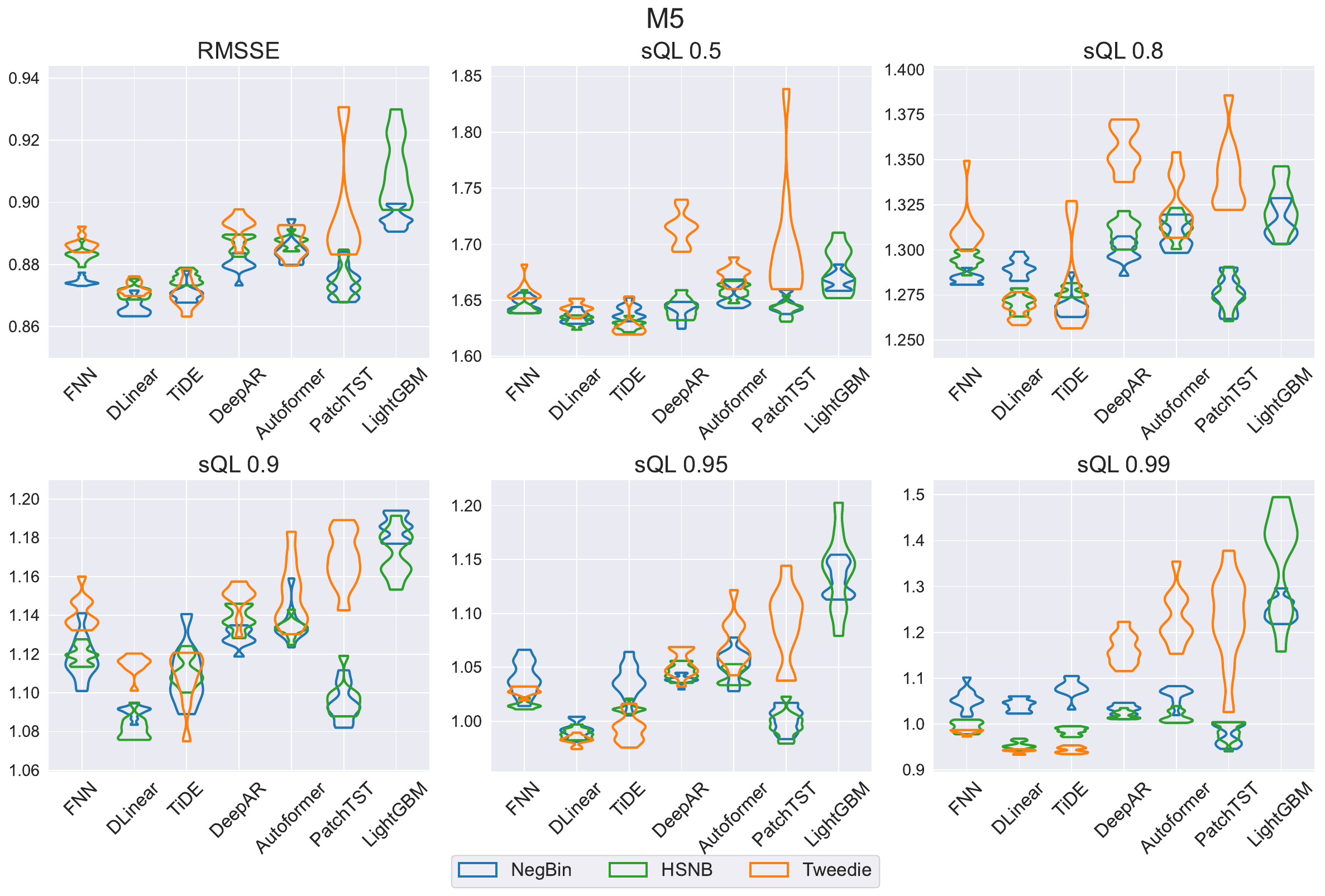}
    \caption{Violin plots of the performance of global models on the M5 data set.}
    \label{fig:all_global_M5}
\end{figure}

\begin{figure}
    \centering
    \includegraphics[width=1\linewidth]{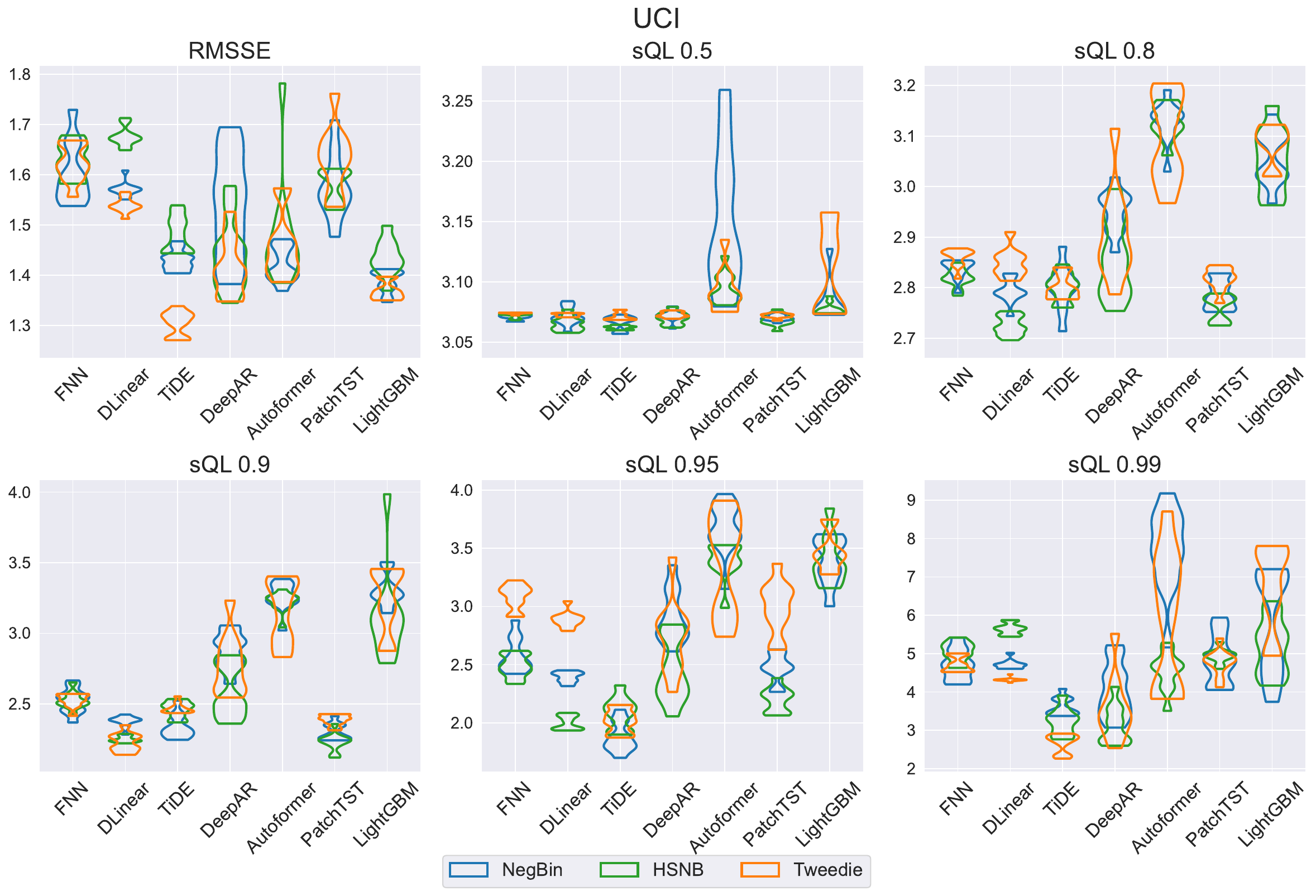}
    \caption{Violin plots of the performance of global models on the UCI data set.}
    \label{fig:all_global_UCI}
\end{figure}

\begin{figure}
    \centering
    \includegraphics[width=1\linewidth]{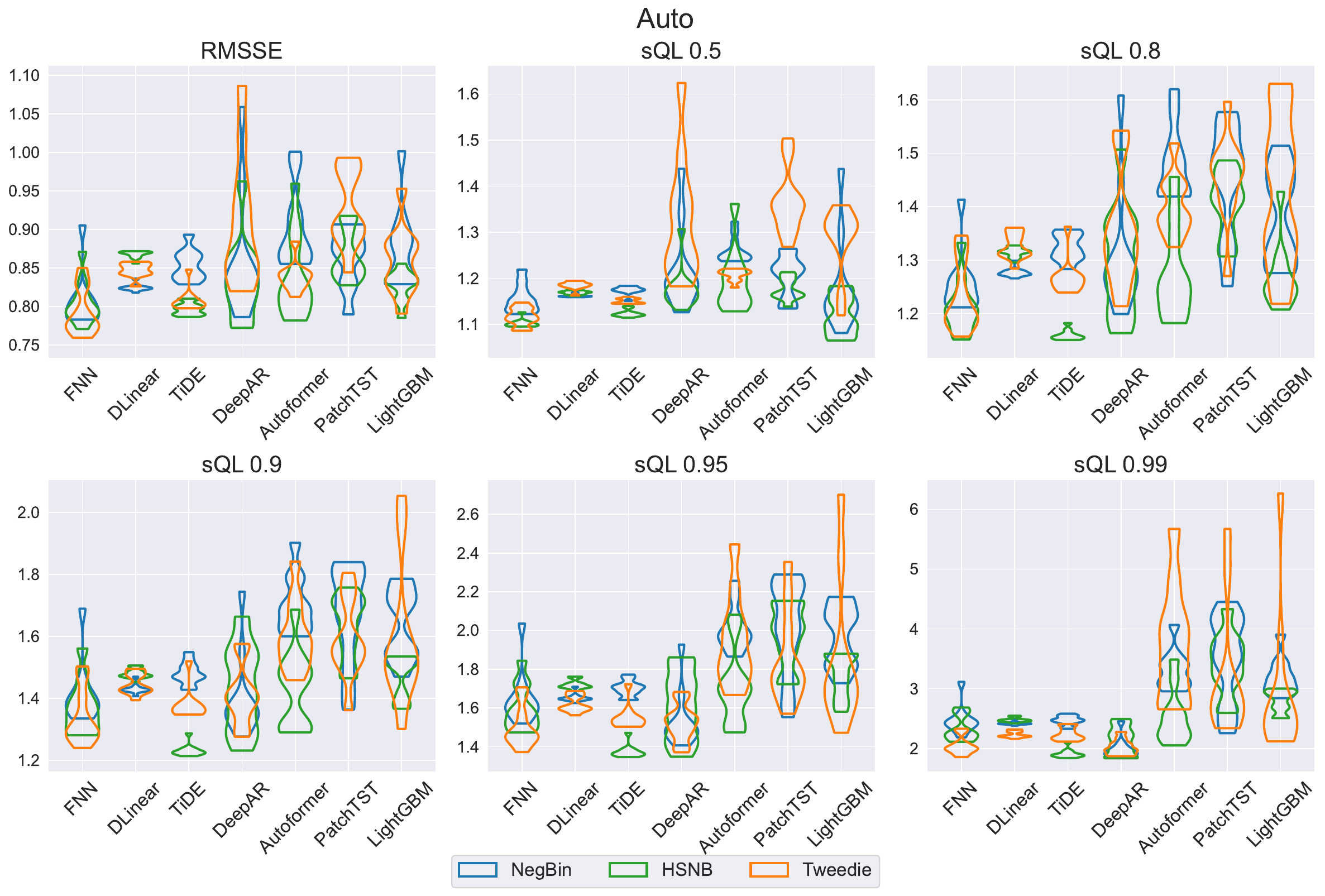}
    \caption{Violin plots of the performance of global models on the Auto data set.}
    \label{fig:all_global_Auto}
\end{figure}

\begin{figure}
    \centering
    \includegraphics[width=1\linewidth]{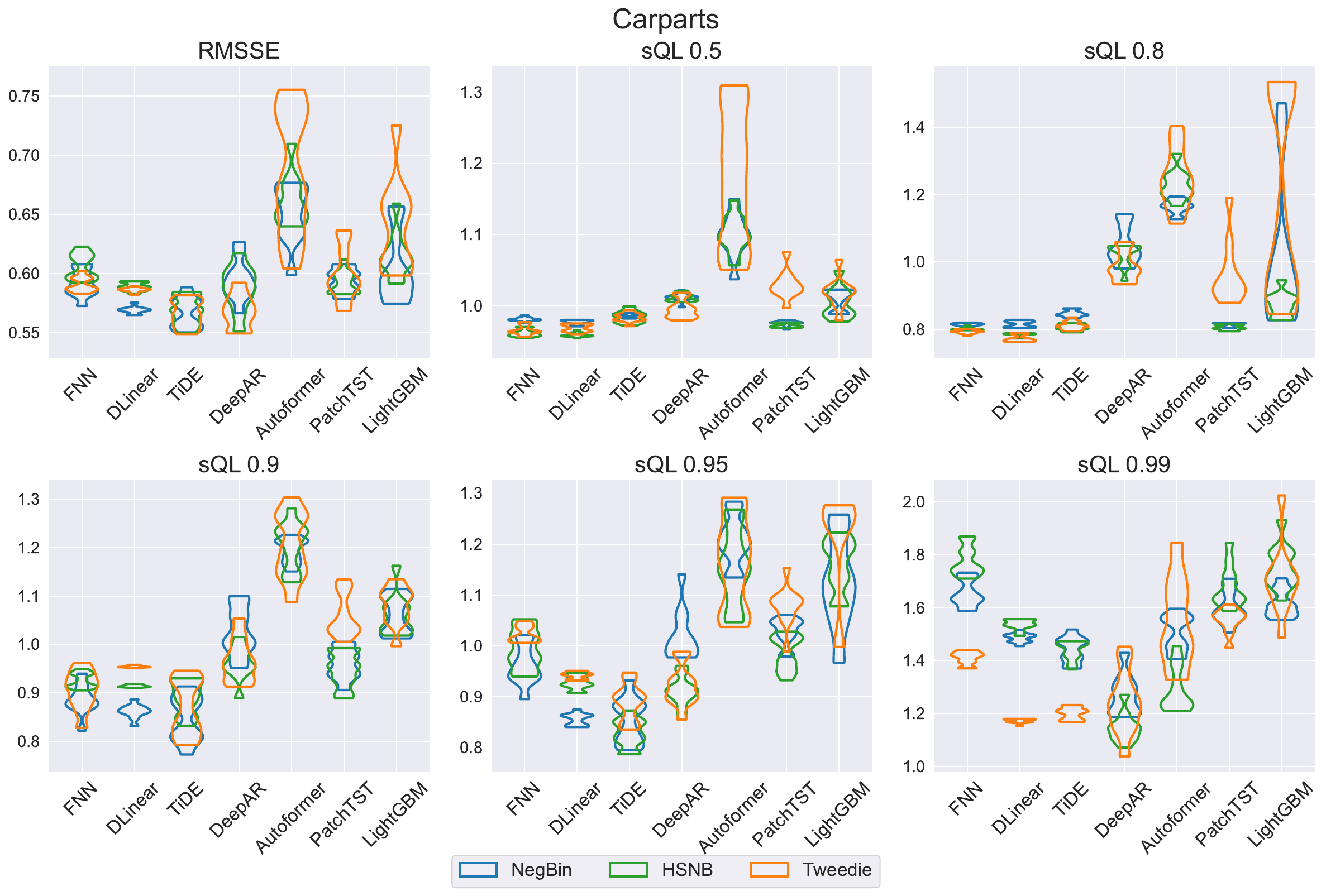}
    \caption{Violin plots of the performance of global models on the Carparts data set.}
    \label{fig:all_global_carparts}
\end{figure}

\begin{figure}
    \centering
    \includegraphics[width=1\linewidth]{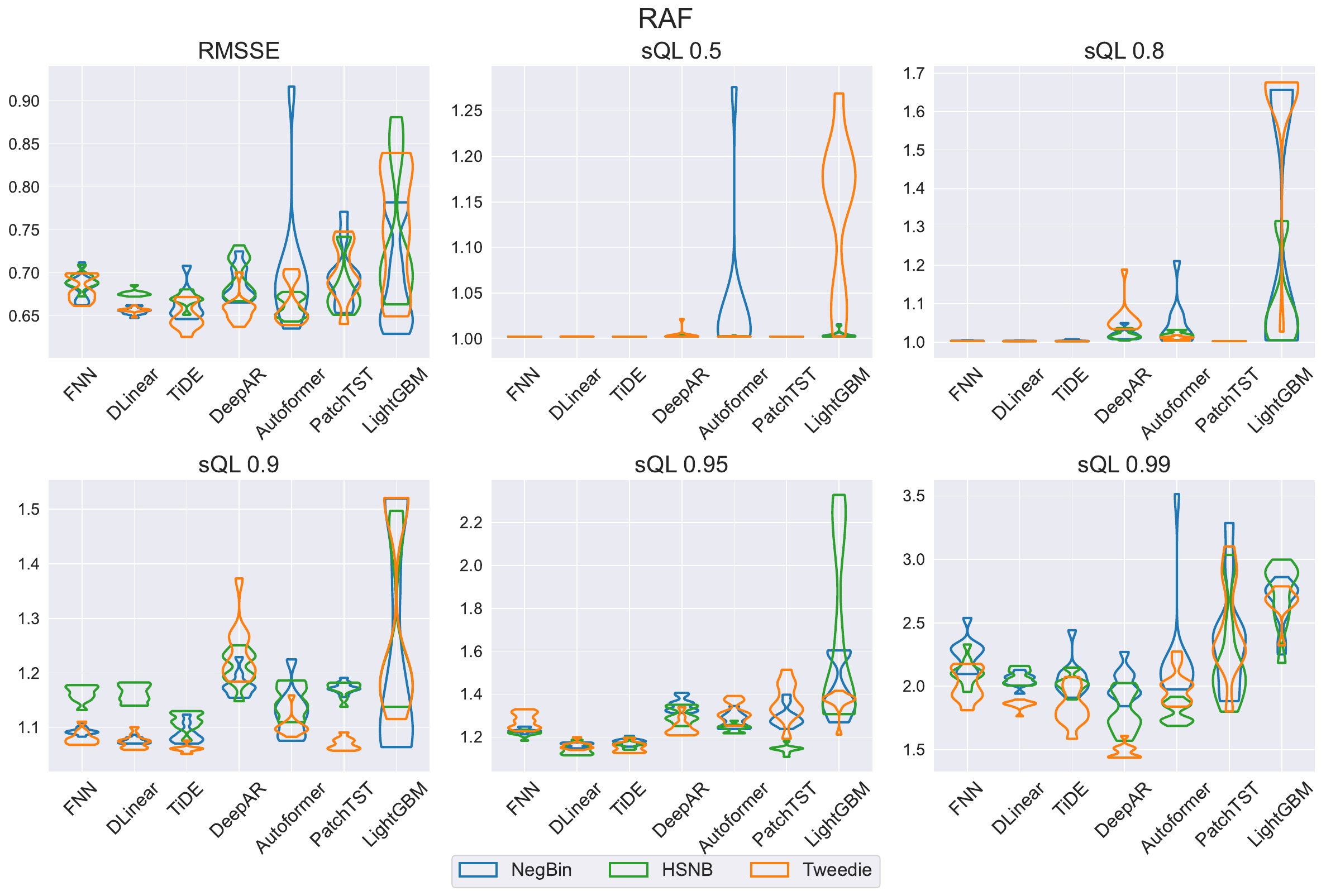}
    \caption{Violin plots of the performance of global models on the RAF data set.}
    \label{fig:all_global_RAF}
\end{figure}

We report in Figs.~\ref{fig:all_global_M5},~\ref{fig:all_global_UCI}.~\ref{fig:all_global_Auto},~\ref{fig:all_global_carparts}, and~\ref{fig:all_global_RAF} the complete scores of the global models on the 5 data sets used in the experiments.

\section{Assessing the distribution head for DLinear model}\label{appendix:dlinear_head}

\begin{table}[!ht]
\centering 
\begin{tabular}{ll|ccccc} 
\toprule 
Metric & Head & M5 & UCI & Auto & Carparts & RAF \\ 
\midrule 

RMSSE & HSNB & \notsign{0.00} & \signpos{0.11} & \signpos{0.04} & \signpos{0.02} & \signpos{0.02} \\ 
RMSSE & Tweedie & \signpos{0.01} & \notsign{-0.03} & \signpos{0.02} & \signpos{0.02} & \notsign{0.00} \\ 
\midrule

sQL$_{0.5}$ & HSNB & \notsign{-0.00} & \notsign{-0.01} & \notsign{0.01} & \signneg{-0.02} & \notsign{-0.00} \\ 
sQL$_{0.5}$ & Tweedie & \signpos{0.01} & \notsign{0.00} & \signpos{0.02} & \notsign{-0.01} & \notsign{0.00} \\ 
\midrule

sQL$_{0.8}$ & HSNB & \signneg{-0.02} & \signneg{-0.07} & \notsign{0.04} & \signneg{-0.03} & \notsign{-0.00} \\ 
sQL$_{0.8}$ & Tweedie & \signneg{-0.02} & \signpos{0.06} & \signpos{0.05} & \signneg{-0.04} & \notsign{-0.00} \\ 
\midrule

sQL$_{0.9}$ & HSNB & \signneg{-0.01} & \signneg{-0.12} & \notsign{0.05} & \signpos{0.05} & \signpos{0.09} \\ 
sQL$_{0.9}$ & Tweedie & \signpos{0.02} & \signneg{-0.15} & \notsign{0.02} & \signpos{0.09} & \notsign{-0.00} \\ 
\midrule

sQL$_{0.95}$ & HSNB & \notsign{-0.00} & \signneg{-0.39} & \signpos{0.07} & \signpos{0.07} & \signneg{-0.02} \\ 
sQL$_{0.95}$ & Tweedie & \signneg{-0.01} & \signpos{0.49} & \signneg{-0.03} & \signpos{0.09} & \notsign{-0.00} \\ 
\midrule

sQL$_{0.99}$ & HSNB & \signneg{-0.09} & \signpos{0.92} & \notsign{0.06} & \signpos{0.05} & \signpos{0.03} \\ 
sQL$_{0.99}$ & Tweedie & \signneg{-0.10} & \signneg{-0.40} & \signneg{-0.19} & \signneg{-0.31} & \signneg{-0.19} \\ 

\bottomrule 
\end{tabular}
\caption{Coefficients of the ANOVA, run on the results of DLinear models. The predictors are given by the interaction between metric and distribution head. The intercept contains RMSSE among metrics, and negative binomial among heads. Coefficients in red and black are statistically significant, and represent respectively an improvement and a worsening compared to the baseline. Coefficients in grey are not statistically significant. We do not report the intercept and coefficients related to the negative binomial baseline, as they are not interpreted in our analysis.}
\label{tab:DLinear_Anova}
\end{table}

Tab.~\ref{tab:DLinear_Anova} shows the same analysis carried out in Sec.~\ref{subsec:head_selection}: in this case however, the DLinear model is considered.

The results confirm that the Tweedie distribution head has generally a better performance on the last percentile (0.99). At less extreme quantiles, however, we observe that it outperforms the negative binomial distribution just as often as it is outperformed by it: the same applies to the HSNB, which has a slight advantage over the negative binomial distribution only at quantile 0.8.

On RMSSE, we observe that the negative binomial performs better than its competitors. In general, these results do not provide evidence that one distribution is preferable to others (contrary to what was seen in Section~\ref{subsec:head_selection} with TiDE): given the inconsistency of results across data sets and metrics, we recommend guiding the choice via cross-validation based on the most important metric at hand.

\end{appendices}

\end{document}